\documentclass[letterpaper, 10 pt, conference]{ieeeconf}  

\IEEEoverridecommandlockouts                              

\usepackage{times}
\usepackage{multicol}
\usepackage[bookmarks=true]{hyperref}
\usepackage{graphics} 
\usepackage{epsfig} 
\usepackage{mathrsfs}
\usepackage[font=small,labelfont=bf]{caption}
\usepackage{times} 
\usepackage{amsmath} 
\usepackage{amssymb}  
\usepackage{algpseudocode}
\usepackage[ruled,vlined,linesnumbered]{algorithm2e}
\usepackage{dblfloatfix}    
\usepackage{optidef}
\usepackage{booktabs}
\usepackage{amsthm}
\usepackage{svg}
\usepackage{cite}
\usepackage{hyperref}
\usepackage{dsfont}
\usepackage{float}

\newtheorem{theorem}{Theorem}
\newtheorem{lemma}{Lemma}
\newtheorem{example}{Example}

\newtheorem{definition}{Definition}

\hypersetup{
    colorlinks=true,
    linkcolor=blue,
    filecolor=green,      
    urlcolor=black,
}

\include{pang_commands}

\renewcommand{\baselinestretch}{0.975}

\title{\LARGE \bf
Bundled Gradients through Contact via Randomized Smoothing}
\author{H.J. Terry Suh$^{1*}$, Tao Pang$^{1*}$, and Russ Tedrake$^{1}$
\thanks{*These authors contributed equally to this work.}
\thanks{*This work was funded by NSF Award No.EFRI-1830901, Navy ONR Award No. N00014-18-1-2210, and Lincoln Lab PO No. 7000470769.}%
\thanks{$^{1}$Massachusetts Institute of Technology, Cambridge, Massachusetts 02139, USA
        {\tt\small \{hjsuh, pangtao, russt\}@mit.edu} }  
}

\makeatletter
\IEEEtriggercmd{\reset@font\normalfont\fontsize{6.8pt}{8.00pt}\selectfont}
\makeatother
\IEEEtriggeratref{1}

\begin{document}

\maketitle
\thispagestyle{empty}
\pagestyle{empty}

\begin{abstract}
The empirical success of derivative-free methods in reinforcement learning for planning through contact seems at odds with the perceived fragility of classical gradient-based optimization methods in these domains. What is causing this gap, and how might we use the answer to improve gradient-based methods? We believe a stochastic formulation of dynamics is one  crucial ingredient. We use tools from randomized smoothing to analyze sampling-based approximations of the gradient, and formalize such approximations through the bundled gradient. We show that using the bundled gradient in lieu of the gradient mitigates fast-changing gradients of non-smooth contact dynamics modeled by the implicit time-stepping, or the penalty method. Finally, we apply the bundled gradient to optimal control using iterative MPC, introducing a novel algorithm which improves convergence over using exact gradients. Combining our algorithm with a convex implicit time-stepping formulation of contact, we show that we can tractably tackle planning-through-contact problems in manipulation.

\end{abstract}

\section{Introduction}

Robots interact with their environments primarily through contact, and planning of such systems (colloquially named \textit{planning through contact}) is a long-standing challenge in manipulation and locomotion, especially in the absence of fixed mode sequences. Neither the tightly coupled problems of \textit{modeling} the contact dynamics nor the \textit{planning} that uses the prescribed model have straightforward answers.

A large body of previous work in planning through contact \cite{posa,contactinvariantoptimization,add,carpentierddp,francois,kent} has been based on gradient-based optimizers. Combining different contact models, optimization methods for planning, and a variety of relaxation schemes, these methods have shown reasonable success. However, they still tend to remain brittle, and their success depends heavily on hyperparameters and the quality of initial guesses. When they fail, they can be notoriously hard to debug.

In another direction, recent works in Reinforcement Learning (RL) have shown impressive results in tackling problems with contact dynamics \cite{fu2016oneshot,nagabandi2019deep,heess2017emergence}. The quality of the solutions obtained via RL is surprising, despite the fact that no underlying structure was used. This suggests that improvements can be made in time and sample complexity. More importantly; however, the fact that we do not understand the gap between the success of RL methods and the struggle of more classical gradient-based methods in planning through contact, is unsatisfying. What are the key improvements that RL made but the classical methods failed to consider? We believe that the answer lies in how RL fundamentally considers a stochastic formulation, as opposed to gradient-based methods that have largely been deterministic.

\begin{figure}[t!]
\centering\includegraphics[width = 0.5\textwidth]{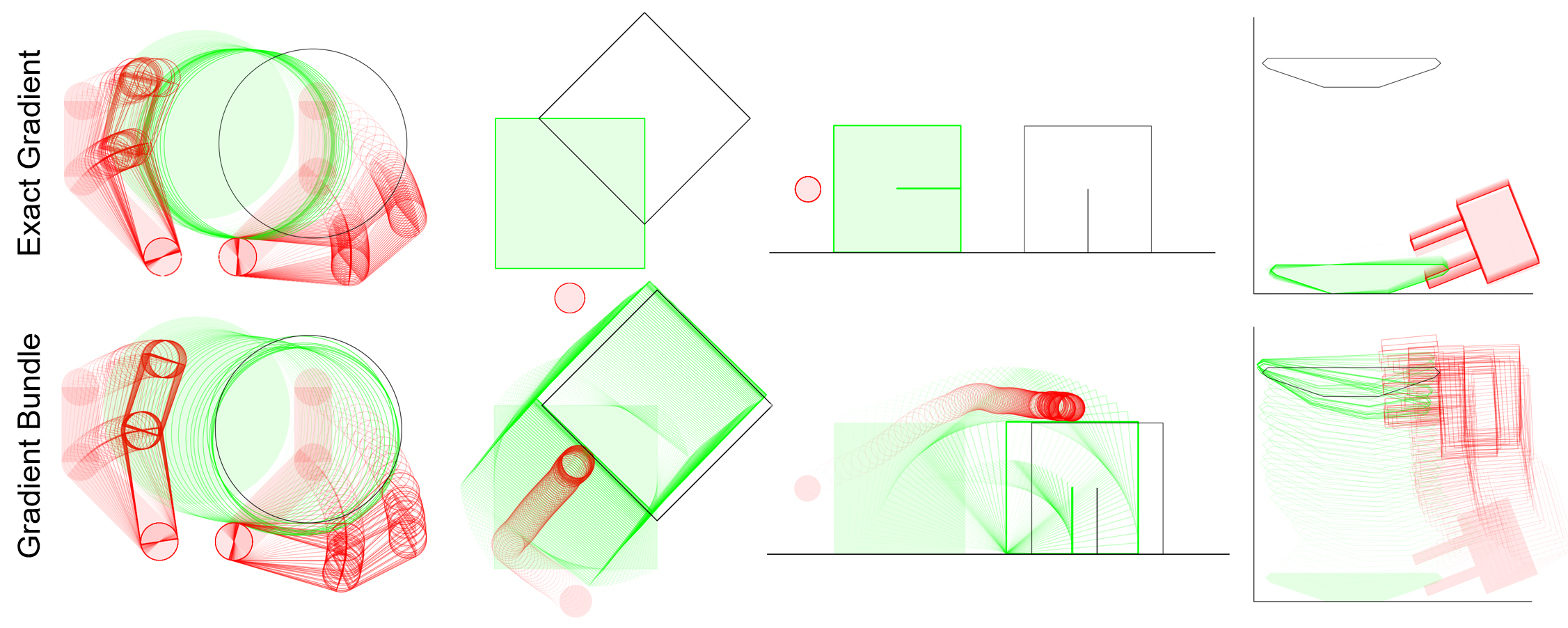}
\caption{Planning with exact gradients (top) vs. the bundled gradient (bottom). Actuated objects drawn in red, unactuated objects drawn in green, and the goal drawn in black. From left: planar hand, planar pushing, box pivoting, and plate pickup task (\cite{florence2019selfsupervised}).}
\label{fig:banner}
\vskip -0.25 true in
\end{figure}

If this is true, what would it mean to take a probabilistic formulation of contact dynamics, and how would it help the planning problem? Can we learn from the answers and combine stochasticity with gradient-based approaches to obtain better convergence behavior? We aim to answer these questions through the lens of \textit{randomized smoothing} \cite{gradientsampling,duchi1, duchi2}. Results from non-smooth optimization \cite{gradientsampling} tell us that the fast-changing gradients of non-smooth problems destabilize gradient-based optimizers such as gradient descent \cite{yuxinchen}. By considering a probabilistic approach to contact, we aim to alleviate flatness of gradients, as well as stabilize fast-changing gradients of contact dynamics to produce a more globally informative direction of improvement. 

We formalize gradients taken through randomized smoothing as the \textit{bundled gradient}, and show that it has benefits over exact gradients of contact models, such as overcoming non-smoothness, alleviating nonconvexity, and supporting a zero-order variant. The bundled gradient can replace the exact gradient in optimization problems involving contact dynamics, including optimal control, policy search, state estimation, and system identification. In our work, we apply bundled gradients on the standard iterative Model Predictive Control (iMPC) algorithm to address planning-through-contact problems, and show that it leads to better convergence behavior.

In addition, randomized smoothing allows obtaining the bundled gradient of contact dynamics that are defined through implicit optimization problems \cite{stewarttrinkle,anitescupotra,carpentierddp}, that are often preferred over penalty methods \cite{huntcrossley} for its ability to take longer timesteps. To demonstrate this advantage, we utilize our method on a quasi-dynamic simulator that uses an implicit time-stepping scheme \cite{Pang2020ACQ} and show that we can tractably tackle robotic manipulation problems.

\textbf{Contributions}. We formalize the notion of the bundled gradient by introducing randomized smoothing, which we then apply to penalty and implicit time-stepping models of contact dynamics. We replace the exact gradients of dynamics with the bundled gradient in iMPC, showing that we can more robustly solve manipulation problems.

\section{Literature Review}
\subsection{Differentiable Modeling of Contact}
We mainly address two classes of approaches for modeling contact: the penalty method \cite{huntcrossley}, and the implicit time-stepping method \cite{stewarttrinkle,anitescupotra}. Although other classes of models exist (e.g. hybrid dynamics simulation through event-detection and impacts \cite{brianmirtich}), they have yet to scale to simulating the complex geometry and dynamics of manipulation.

Works that utilize the penalty method \cite{add,tobia2,terryfelix} approximate contact behavior at each timestep with forces from stiff spring and dampers. While the method can approximate a wide range of contact behavior such as impact, the dynamics often results in stiff ordinary differential equations (ODE), which require small timesteps to solve numerically. In addition, although penalty methods often have straightforward expressions for the derivative \cite{add}, the computation of long-horizon gradients may be expensive due to large number of timesteps required.

On the other hand, implicit time-stepping methods \cite{posa,variablesmoothing,carpentierddp} can take larger timesteps without sacrificing integration accuracy, but integration requires solving a Linear Complementarity Problem (LCP) \cite{cottle2009linear}, whose per-step computation is harder and solutions may not be unique. Nevertheless, the LCP can be converted to a convex Quadratic Program (QP) by relaxing the Coulomb friction constraints \cite{anitescu2006optimization, convexsmoothcontactmodel, Pang2020ACQ}. The implicit models also allow differentiation by computing the derivatives of the Karush-Kuhn-Tucker (KKT) conditions of the QP using the implicit function theorem \cite{boot1963sensitivity}.  

\subsection{Smooth Approximations of Contact for Planning}

As solving the non-smooth problem directly (e.g. via mixed-integer programming \cite{tobia,francois}) is difficult, many existing works consider solving a smooth approximation of the problem. For instance, penalty-based methods make explicit smooth approximations of the contact dynamics \cite{add, tobia2,terryfelix}. While the behavior of such approximations have intuitive interpretations at a dynamics-level, the choice of functions that are used is arbitrary. In addition, it is not clear how to apply such explicit smoothing to implicit dynamics whose result is defined by an optimization problem. 

Other works \cite{variablesmoothing, posa, scott, convexsmoothcontactmodel, contactinvariantoptimization, stochasticcomplementarity} do not directly smooth out the dynamics, but find an explicit relaxation of constraints that implicitly define the dynamics. While these works are often better connected to the underlying optimization, the interpretation of the resulting dynamics can be unintuitive which makes debugging cumbersome. 

Our philosophy behind why stochasticity is necessary for planning is similar to the argument presented in \cite{stochasticcomplementarity}. However, unlike \cite{stochasticcomplementarity} which explicitly uses ``rounded" complementarity constraints and interprets it as minimizing expected constraint violation, randomized smoothing is fully implicit \cite{duchi1,duchi2}, in the sense that we do not require explicit smoothing formulas for dynamics, nor the constraints that define them. This allows us to make smooth approximations to a more general class of models. In addition, the bundled dynamics from randomized smoothing is more physically-interpretable than numerical smoothing strategies that ``sand off" the sharp corners of non-smooth constraints.

\subsection{Randomized Smoothing for Non-smooth Optimization}
The destabilization of gradient descent by non-smooth functions was observed by early works such as Wolfe's example \cite{yuxinchen,clarke}. Stabilized gradient descent through \textit{gradient sampling} was introduced in \cite{gradientsampling,subdifferentials,sqpgs}, which proposed to approximate an instance of a subgradient in the convex hull of sampled gradients around the current iterate. Based on these works, an optimal convergence guarantee was provided in \cite{duchi1} where the term \textit{randomized smoothing} was introduced, and a zero-order variant was presented in \cite{duchi2}.

Our work strongly builds upon existing works, but we analyze the consequences of randomized smoothing in the presence of contacts and aim to use the bundled gradient for optimal control. In addition, we show potential pitfalls of gradient sampling in the presence of dynamics of discontinuities, which has not been addressed by previous works.  

\section{Randomized Smoothing}
\label{sec:randomizedsmoothing}

Motivated by \cite{gradientsampling, duchi1}, we first introduce the idea of randomized smoothing in the context of gradient descent.

\subsection{The Bundled Gradient and the Bundled Objective}
Let $f(x)$ be a non-smooth objective function that is differentiable almost everywhere, and denote the gradient as $\nabla f(x)$. While the gradient suggests a local direction of descent, it is myopic. For smooth non-convex functions, gradient descent could converge to bad local minima. For non-smooth functions, the gradient may be subject to large jumps, stalling the descent on the cost function \cite{clarke, yuxinchen} and compromising standard  guarantees of convergence. To alleviate the problems caused by the locality of the gradient, we can aggregate gradient information in some neighborhood, which we formalize with the term \textit{bundled gradient}.


\begin{definition}
    \normalfont \textbf{Bundled Gradient.} Let $\mu\in L^1$ be a symmetric probability density. Then, the \textit{first-order bundled gradient} is defined as 
    \vskip -0.1 true in
    \begin{equation}
        \small
        \bar{\nabla} f(x) := \int \nabla f(x+w)\mu(w)dw = \mathbb{E}_{w\sim\mu}[\nabla f(x+w)].
    \end{equation}
\end{definition}

The bundled gradient can be thought of as a convolution of $\nabla f(x)$ and the kernel $\mu$. Further insight on the bundled gradient can be obtained by defining the \textit{bundled objective}.

\begin{definition}
    \normalfont \textbf{Bundled Objective.} The bundled objective is defined as
    \begin{equation}
        \small
        \bar{f}(x) := \int f(x+w)\mu(w)dw = \mathbb{E}_{w\sim \mu}[f(x+w)].
    \end{equation}
\end{definition}

\begin{lemma}
    \normalfont The first-order bundled gradient $\bar{\nabla}f(x)$ is the exact gradient of the bundled objective $\bar{f}(x)$, $\bar{\nabla}f(x)=\nabla \bar{f}(x)$.
    \begin{proof}
        Exchange the expectation $\mathbb{E}_w$ and the derivative $\nabla_x$ using the Dominated Convergence Theorem.
    \end{proof}
    \label{lemma:dct}
\end{lemma}

This suggests that using the bundled gradient in place of the exact one lets us implicitly operate on a different objective function defined by the bundled objective. The bundled objective has an intuitive explanation as result of filtering via convolution, such as Gaussian smoothing. Such smoothing schemes alleviate non-convexity by smoothing out local minima, and allows tackling non-smooth problems by stabilizing gradient descent \cite{gradientsampling}. Note that the bundled gradient is purposely biased to produce more global effects, and thus will not converge to a stationary point of the original objective, but rather that of the bundled objective. This can be mitigated by iteratively decreasing the variance $\sigma(\mu)$, as done in typical stochastic approximation schemes \cite{robbinsmonro,kieferwolfowitz}.

\subsection{Zero-Order Variation}
If $\nabla f$ is not available or costly to compute, one may instead compute an expected value of finite-differences (which we denote by $\nabla_0 f(x)$) to use as a direction of descent \cite{duchi2}. We call this object the \textit{zero-order bundled gradient}, 
\begin{equation}
    \small
    \big[\bar{\nabla}_0 f(x)\big]_j := \mathbb{E}_{w\sim\mu}[\nabla_0 f(x)]_j = \mathbb{E}_{w\sim\mu} \bigg[\frac{f(x+w_j)-f(x)}{w_j}\bigg],
    \label{eq:zeroorderbundledgradient}
\end{equation}
where subscript $j$ denotes the $j^{\text{th}}$ component of the vector. Convergence of such schemes are analyzed in Simultaneous Perturbation Stochastic Approximation (SPSA) \cite{spsa}.


\begin{example}
    \normalfont\textbf{Function with many Local Minina}: Let 
    \begin{equation}
        f(x)=x^2 + 0.1\sin(20x),
    \end{equation}
    which has many bad local minima. Under a suitable choice for the Gaussian kernel $\mu(w)=\mathcal{N}(w;0,\sigma)$, this function can be smoothed out to be convex, or gradient dominant \cite{gradientdominance}.
    \begin{figure}[thpb]
    \vskip -0.1 true in
	\centering\includegraphics[width = 0.45\textwidth]{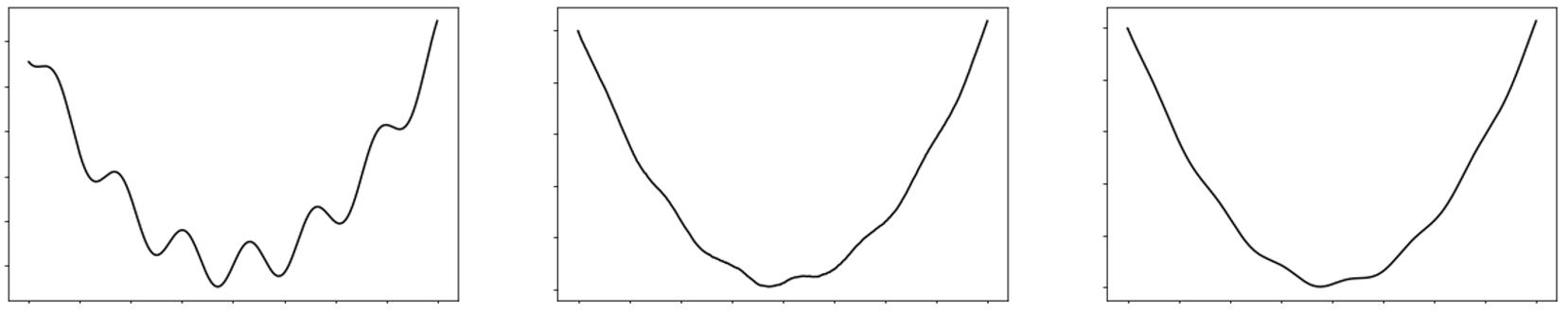}
    \caption{Left: Objective $f(x)$. Center: Bundled objective reconstructed from first-order bundled gradient. Right: Bundled objective reconstructed from zero-order bundled gradient}
    \label{fig:localminima}
	\vskip -0.2 true in
    \end{figure}
\end{example}

\subsection{Monte-Carlo Integration for the bundled gradient}
Symbolic computation of the bundled gradient is prohibitive, and quadrature methods do not scale well with dimension. Thus, we propose to use Monte-Carlo integration, which can scale to high dimensions under suitable conditions on the variance of the objective. This yields the gradient sampling algorithm \cite{gradientsampling} and its zero order variant \cite{duchi2},
\vskip -0.2 true in
\begin{equation}
    \small
    \begin{aligned}
    \hat{\nabla} f(x) & \approx \frac{1}{N} \sum_i \nabla f(x+w_i) \quad w_i \sim \mu(w) \\
    \hat{\nabla}_0 f(x)_j  & \approx \frac{1}{N} \sum_i \bigg(\frac{f(x+w_{ij})-f(x)}{w_{ij}}\bigg) \quad w_i \sim \mu(w),
    \end{aligned}
\end{equation}
where $i$ is the index of the sample, and $j$ is defined as (\ref{eq:zeroorderbundledgradient}).

\begin{lemma}
    \normalfont \textbf{Law of Large Numbers}. If $f$ has no discontinuous jumps and $\nabla f, \nabla f_0 \in L^1$, then $\hat\nabla f(x)\xrightarrow[]{a.s.} \bar{\nabla}f(x)$ and $\hat{\nabla}_0 f(x)\xrightarrow[]{a.s}\bar{\nabla}_0 f(x)$, where $a.s.$ stands for almost surely. 
    \label{lemma:lln}
\end{lemma}

However, one must be very careful with this Monte Carlo approximation when $f$ has discontinuous jumps, due to the dirac-delta that appears in the derivative $\nabla f$ \cite{teg}. We formalize this through the following theorem:

\begin{theorem} \normalfont \textbf{Sampling Gradients of Discontinuities.}
    Let $f(x)$ be a function differentiable almost everywhere with at least one jump discontinuity; $\exists x' \;\text{s.t. } \lim_{x\rightarrow x'^-} f(x)\neq \lim_{x\rightarrow x'^+} f(x)$. Then, the Monte-Carlo estimate of the first-order bundled gradient almost surely fails to converge to the gradient of the bundled dynamics, $\bar{\nabla} f(x) \neq^{\text{a.s.}} \nabla \bar{f}(x).$

    \label{theorem:discontinuity}
    \begin{proof}
        Sampling from the dirac-delta is ill-defined, as it does not evaluate pointwise. In practice, however, such an attempt would always lead to $0$ almost surely. Without loss of generality, let $x'=0$. Then, $\exists f_L,f_R$ s.t. $\small c = \small f_R(0) - \small f_L(0)\neq 0$:
        \begin{equation}
            \small
            \begin{aligned}
                f(x) & = f_R(x)\mathds{1}_{x\geq 0}(x) + f_L(x) \mathds{1}_{x <0}(x) \\
                \nabla f(x) & =\nabla f_R(x)\mathds{1}_{x>0}(x) + \nabla f_L(x)\mathds{1}_{x<0}(x) + c \delta(x).
            \end{aligned}
        \end{equation}
        where $\mathds{1}_A(x)$ is the indicator function for set $A$.
        \begin{equation}
            \small
            \begin{aligned}
                \nabla \bar{f}(x) & = \int \nabla f(x+w)\mu(w)dw \\ 
                & = c\mu(x) + \int \big(\nabla f(x+w) - c\delta(x+w)\big)\mu(w)dw.
            \end{aligned}
        \end{equation}
        Given that samples drawn from $w_i\sim\mu(w)$ have zero probability of satisfying $x+w_i$, 
        \begin{equation}
            \small
            \begin{aligned}
            \hat{\nabla}f(x)& \xrightarrow[]{a.s.} \int (\nabla f(x+w) - c\delta(x+w))\mu(w)dw, \\
            \nabla \bar{f}(x) - \hat{\nabla} f(x) & \xrightarrow[]{a.s.} c\mu(x).
            \end{aligned}
        \end{equation}
        which completes the proof.
    \end{proof}
\end{theorem}

On the other hand, the zero-order variant does not suffer from this by using finite intervals for approximation.We illustrate an extreme case of Theorem \ref{theorem:discontinuity} in Example \ref{ex:heaviside}.

\begin{example}
    \label{ex:heaviside}
    \normalfont\textbf{Heaviside Function}: Let 
    \begin{equation}
        f(x)=
            1, \text{ if } x \geq 0, f(x)=0, \text{ if } x < 0 .
    \end{equation}
    Then, $\nabla f(x)=\delta(x)$, and the approximated bundled gradient $\bar{\nabla} f(x+w_i)=0$ almost surely. However, the analytical computation of the bundled objective with a Gaussian $\mu$ would result in $\bar{f}(x)=\text{erf}(x)$, whose derivative is non-zero at $x=0$. Thus $\hat{\nabla}f(x)\neq \nabla \bar{f}(x)$.
    
    \begin{figure}[thpb]
	\centering\includegraphics[width = 0.45\textwidth]{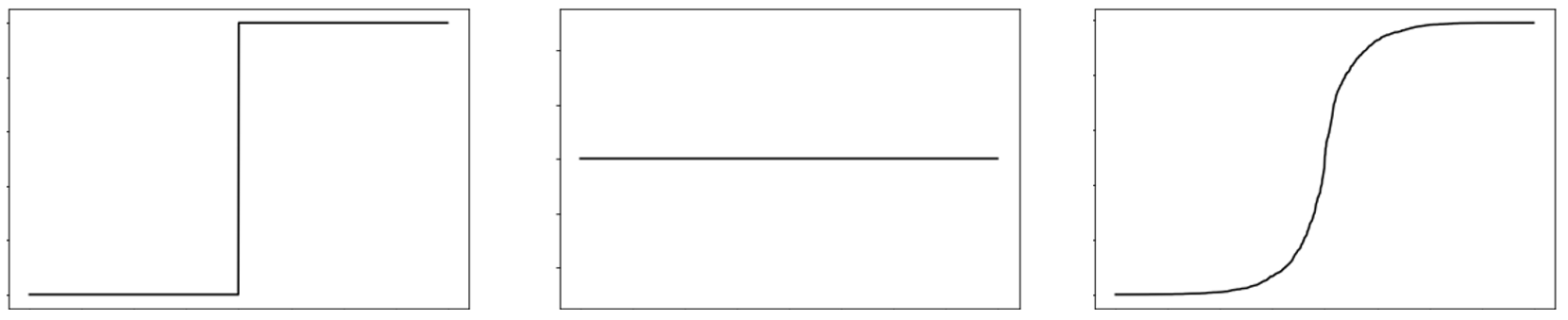}
    \caption{Left: Objective $f(x)$. Center: Bundled objective reconstructed from first-order bundled gradient. Right: Bundled objective reconstructed from zero-order bundled gradient}
	\label{fig:heaviside}
	\vskip -0.2 true in
    \end{figure}
\end{example}

\subsection{Bundled Jacobian and Bundled Dynamics}
We extend the concept of randomized smoothing to dynamical systems with the \textit{bundled Jacobian} and the \textit{bundled dynamics}. Let $f$ be a discrete-time dynamical system with state $x_t\in\mathbb{R}^n$, and input $u_t\in\mathbb{R}^m$,
\begin{equation}
\label{eq:generic_discrete_time_dynamics}
\small
    x_{t+1} = f(x_t, u_t).
\end{equation}

Recall that the linearization of the dynamical system $f$ around the nominal trajectory $\{\bar{x}_t,\bar{u}_t\}^T_{t=0}$ can be formed by taking a first-order Taylor expansion: 
\begin{equation}
\small
    \label{eq:linearization}
    \begin{aligned}
    x_{t+1} & = \underbrace{\frac{\partial f}{\partial x}\bigg|_{x=\bar{x}_t,u=\bar{u}_t}}_{\mathbf{A}_t} x_t + \underbrace{\frac{\partial f}{\partial u}\bigg|_{x=\bar{x}_t,u=\bar{u}_t}}_{\mathbf{B}_t} u_t + c_t, \\
    c_t & = f(\bar{x}_t, \bar{u}_t) - \mathbf{A}_t \bar{x}_t - \mathbf{B}_t \bar{u}_t,
    \end{aligned}
\end{equation}
which provides a local and myopic approximation of $f$. We apply randomized smoothing to the Jacobian to obtain a more informative linear approximation.

\begin{definition}
    \normalfont\textbf{The Bundled Jacobian} Let $\mu\in L^1$ be a multivariate symmetric probability density with arguments $w\in\mathbb{R}^m,v\in\mathbb{R}^n$. Then, the bundled Jacobian is defined as a convolution between $\mu(w,v)$ and the Jacobian of $f$:
    \small
    \begin{equation}
        \begin{aligned}
            \bar{\mathbf{A}}_t = \mathbb{E}_{w,v}\bigg[\frac{\partial f}{\partial x}\bigg|_{(\bar{x}_t+w, \bar{u}_t + v)}\bigg],
            \quad\bar{\mathbf{B}}_t = \mathbb{E}_{w,v}\bigg[\frac{\partial f}{\partial u}\bigg|_{(\bar{x}_t+w, \bar{u}_t + v)}\bigg]. \\
        \end{aligned}
    \end{equation}
\end{definition}

\begin{definition}

    \normalfont\textbf{The Bundled Dynamics} The Bundled dynamics are defined as a multivariate convolution between the original dynamics and the probability density $\mu$:
    \begin{equation}
        \small
        \begin{aligned}
            \bar{f}(x_t,u_t) 
                             & = \mathbb{E}_{w,v}\bigg[f(x_t + w, u_t + v)\bigg].
        \end{aligned}
    \end{equation}
\end{definition}
\begin{lemma}
    \normalfont The system described by the bundled Jacobian $\bar{\mathbf{A}}_t,\bar{\mathbf{B}}_t$ is the linearization of the bundled dynamics $\bar{f}(x_t,u_t)$.  
    \begin{proof}
        Identical to Lemma \ref{lemma:dct}.
    \end{proof}
\end{lemma}

The above theorem establishes that using the bundled Jacobian, as opposed to the exact linearization of the system, is equivalent to implicitly operating on a smoother version of the dynamics that has been convolved with $\mu$. 

\subsection{Monte-Carlo Approximation and Zero-Order Variant}
We propose to approximate the bundled Jacobian using Monte-carlo integration, where $w_i,v_i\sim \mu(w,v)$:
\begin{equation}
    \footnotesize
    \begin{aligned}
    \hat{\mathbf{A}}_t & \approx \frac{1}{N} \sum^N_{i=1} \bigg(\frac{\partial f}{\partial x}\bigg|_{\bar{x}_t+w_i,\bar{u}_t+v_i}\bigg), \quad \hat{\mathbf{B}}_t \approx \frac{1}{N} \sum^N_{i=1} \bigg(\frac{\partial f}{\partial u}\bigg|_{\bar{x}_t+w_i,\bar{u}_t+v_i}\bigg).
    \end{aligned}
    \label{eq:bundlejacobian}
\end{equation}
Finally, we introduce a zero-order variant that takes a least-squares approximation to the local dynamics:

\vskip -0.15 true in
\begin{equation}
    \small
    \begin{aligned}
    \hat{\mathbf{A}}_t^0, \hat{\mathbf{B}}_t^0 & \in \text{argmin}_{\mathbf{A,B}}  \\  &  \sum^N_{i=1}\bigg\|f(\bar{x}_t+w_i, \bar{u}_t+v_i) - f(\bar{x}_t,\bar{u}_t) -\mathbf{A}w_i - \mathbf{B}v_i\bigg\|_2^2.
    \end{aligned}
    \label{eq:zeroorderjacobian}
\end{equation}

\section{Randomized Smoothing of Contact Dynamics}
What does it mean to apply randomized smoothing in the presence of contacts? Intuitively, different samples will encounter different contact modes. In expectation, the effect of sampling is a smoothed behavior where normal and frictional forces are applied at a distance, and the boundaries of stick and slip blur. We analyze this intuition through two classes of contact models: implicit contact defined by complementarity constraints \cite{anitescupotra, stewart2000rigid}, and the penalty method \cite{huntcrossley}. 


\begin{figure*}[!b]
\centering
\includegraphics[width = 0.95\textwidth]{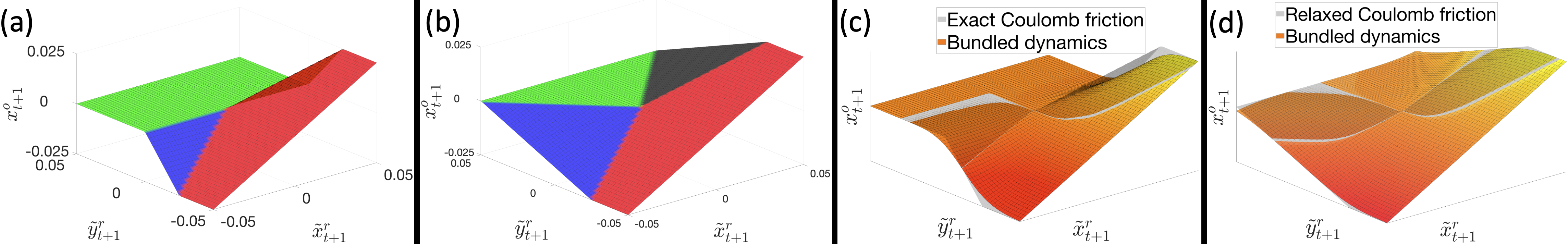}
\caption{Implicit contact dynamics using (\textbf{a}) the exact Coulomb friction model \cite{stewart2000rigid} and (\textbf{b}) the relaxed Coulomb friction model \cite{anitescu2006optimization}. Both (\textbf{a}) and (\textbf{b}) plot the the green box position $x^o_{t+1}$ as a function of the commanded sphere positions $[\tilde{x}^r_{t+1}, \tilde{y}^r_{t+1}]$. Both dynamics are piecewise-linear, where each piece corresponds to a contact mode: green represents separation; blue sliding left, gray sliding right and red sticking. The larger sliding region in (\textbf{b}) is a result of the ``boundary layer" artifact of the relaxed Coulomb friction model.
(\textbf{c}) and (\textbf{d}): bundled dynamics of (\textbf{a}) and (\textbf{b}), respectively, obtained using randomized smoothing. Units in (\textbf{a}) and (\textbf{b}) are in meters.}
\label{fig:friction_dynamics}
\vspace{-0.6cm}
\end{figure*}
\subsection{Contacts Defined by Complementarity Constraints}

Implicit models of contact using complementarity constraints \cite{stewarttrinkle,anitescupotra} are widely used in planning through contact for their ability to stably take bigger time steps \cite{Landry2019BilevelOF, posa}. However, unlike penalty methods which permit explicit smoothing of forces \cite{add,tobia2,terryfelix}, smoothing results of implicit optimization problems is less straightforward; randomized smoothing is unique in that it provides the ability to do so without an explicit relaxation of constraints.

An accurate treatment of multibody dynamics with contact and friction in its full glory requires heavy notation and detracts from the intuition behind the advantage of smoothing. We instead demonstrate the effect of smoothing on contact dynamics through two simple, 2D examples, and refer the reader to \cite{anitescupotra, stewart2000rigid, anitescu2006optimization, Pang2020ACQ} for a more detailed presentation of contact dynamics. 

To simplify the dynamics aspect of the modeling and focus more on contact, we keep the full second-order dynamics for un-actuated objects, but make the assumption that the actuated bodies (robots) are gravity-compensated, PD-controlled and \textit{quasi-static} \cite{pang1996complementarity, Pang2020ACQ}: the robots are always in force equilibrium, with the generalized forces exerted by contact balanced by the virtual spring of the proportional part of the PD controller. The quasi-static assumption reduces the dimension of robot states by half, and is commonly made in systems where inertial forces are negligible compared to contact forces, such as robotic manipulation \cite{francois, chavan2018stable}. Despite these simplifications, our analysis on the effect of contact smoothing does transfer to the full second-order dynamics of the entire robot-object system.

\begin{figure}[h!]
\vspace{-0.5cm}
\centering
\includegraphics[width = 0.42\textwidth]{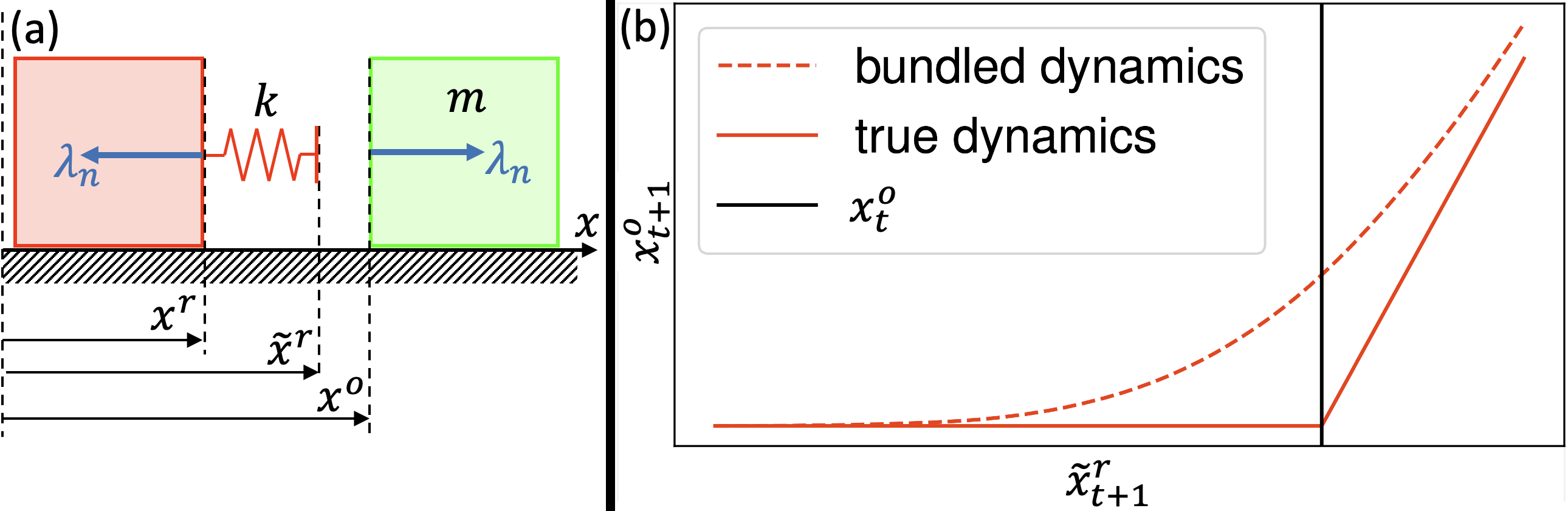}
\caption{(\textbf{a}) A 1D dynamical system consisting of a 1-DOF robot (red box) and an un-actuated object (green box). Both boxes are constrained to slide on a friction surface. (\textbf{b}) The piecewise-linear relationship between the free object position $x^o$ and the commanded robot position $\tilde{x}^r$. Note that there is no contact unless $\tilde{x}^r_{t+1} \geq x^o_{t}$.}
\label{fig:1d_contact}
\vspace{-0.5cm}
\end{figure}

\begin{example}
\normalfont\textbf{Randomized Smoothing of Normal Contact.}
We first consider a 1D system of two objects in Fig. \ref{fig:1d_contact}a, where the boxes can only interact through the normal contact forces. The dynamics of the robot (red box) is described by
\begin{equation}
\label{eq:1d_dynamics:robot}
\small
-\lambda_n + hk (\tilde{x}^r_{t+1} - x^r_{t+1}) = 0,
\end{equation}
where $\lambda_n$ denotes the impulse generated by the normal force during the timestep $h$; $\tilde{x}^r_{t+1} $ and $x^r_{t+1}$ are the robot's \textit{commanded} and actual positions at the next timestep $t+1$. Constraint (\ref{eq:1d_dynamics:robot}) states that the contact impulse $-\lambda_n$ is balanced by the virtual spring impulse $hk (\tilde{x}^a_{t+1} - x^a_{+1})$ from $t$ to $t+1$.

The dynamics of the object (green box) is given by 
\begin{subequations}
\small
\begin{align}
m v^o_{t+1} &= \lambda_n, \label{eq:1d_dynamics:object}\\
x^o_{t+1} &= x^{o}_t + h v^o_{t+1},
\end{align}
\end{subequations}
where (\ref{eq:1d_dynamics:object}) states that the gain in object momentum comes from the normal contact impulse $\lambda_n$, assuming that the object has 0 velocity at the current timestep.

The contact impulse and the distance between the two objects satisfy the following complementarity constraint:
\begin{equation}
\label{eq:1d_dynamics:complementarity}
\small
0 \leq \lambda_n \perp (x^o_{t+1} - x^r_{t+1}) \geq 0,
\end{equation}
where $a \perp b$ for vectors $a$ and $b$ means that $a^\intercal b = 0$. The implication of (\ref{eq:1d_dynamics:complementarity}) is three-fold: (\textbf{i}) the impulse can only push ($\lambda_n \geq 0$); (\textbf{ii}) the two boxes cannot penetrate at $t+1$ ($x^o_{t+1} - x^r_{t+1} \geq 0$); and (\textbf{iii}) non-zero normal impulse exists only if the two objects are in contact ($\lambda_n \perp x^o_{t+1} - x^r_{t+1}$).

The state of the 1D system at $t+1$ can be solved from the implicit contact dynamics (\ref{eq:1d_dynamics:robot}) - (\ref{eq:1d_dynamics:complementarity}). The dynamics consists of two linear pieces, depending on whether the two objects make contact at $t+1$:
\begin{subequations}
\label{eq:1d_dynamics:explicit}
\small
\begin{align}
\begin{bmatrix}
x^o_{t+1} \\
x^r_{t+1} 
\end{bmatrix} 
&=
\begin{bmatrix}
1 & 0 \\
0 & 0
\end{bmatrix}
\begin{bmatrix}
x^o_{t} \\
x^r_{t} 
\end{bmatrix}
+
\begin{bmatrix}
0\\1
\end{bmatrix}
\begin{bmatrix}
\tilde{x}^r_{t+1}
\end{bmatrix}
\; \text{(no contact),} \\
\begin{bmatrix}
x^o_{t+1} \\
x^r_{t+1} 
\end{bmatrix} 
&=
\begin{bmatrix}
\frac{c}{1 + c} & 0 \\
\frac{c}{1 + c} & 0
\end{bmatrix}
\begin{bmatrix}
x^o_{t} \\
x^r_{t} 
\end{bmatrix}
+
\begin{bmatrix}
\frac{1}{1 + c} \\ \frac{1}{1 + c}
\end{bmatrix}
\begin{bmatrix}
\tilde{x}^r_{t+1}
\end{bmatrix}
\; \text{(contact),}
\end{align}
\end{subequations}
where $c = m/(h^2k)$. The explicit dynamics (\ref{eq:1d_dynamics:explicit}), which has the same form as (\ref{eq:linearization}), results in perfect position command being achieved for $x^r_{t+1}$ with no movement for $x^o_{t+1}$ if there is no contact; and $x^r_{t+1}=x^o_{t+1}$ being placed near the commanded position if there is contact. 

As shown in Fig. \ref{fig:1d_contact}b, before making contact, the control input $\tilde{x}^r$ has no effect on $x^o$, producing zero gradients. In contrast, when sampling around $\tilde{x}^r$, some of the samples will hit the un-actuated box, and in expectation, push them away from each other. Thus, the bundled dynamics suggests that $\tilde{x}^r$ should approach $x^o$ to push the box.
\end{example}

\begin{figure}[tp]
\centering
\vskip 0.1 true in
\includegraphics[width = 0.24\textwidth]{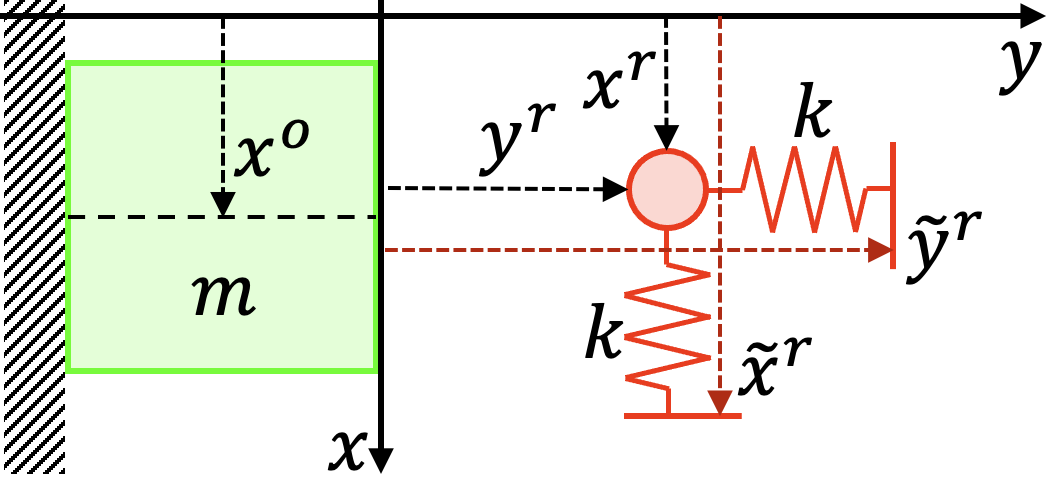}
\caption{A 2D dynamical system consisting of a 2-DOF robot (red sphere) and an un-actuated object (green box). The robot can move freely in the $xy$-plane, whereas the box is constrained to slide on the frictionless surface.}
\label{fig:2d_contact}
\vskip -0.3 true in
\end{figure}

\begin{example}
\normalfont\textbf{Randomized Smoothing of Friction.} In this example, we study the 2D system in Fig. \ref{fig:2d_contact} where the robot (red sphere) interacts with the green box through a frictional contact on the right of the box.

The Coulomb friction model constrains contact force to stay inside the friction cone. In addition, when the contact is sliding, the force needs to stay on the friction cone's boundary and in the opposite direction of the relative sliding velocity. Similar to the normal contact constraint (\ref{eq:1d_dynamics:complementarity}), Coulomb friction can be modeled with additional complementarity constraints \cite{stewarttrinkle}.

Assuming the same quasi-static dynamics for the robot, zero initial velocity for the box, and the initial condition $[x^o_t, x^r_t, y^r_t] = [0, 0, 0]$, it can be shown, by solving the dynamics subject to the Coulomb friction constraints, that the box's position at the next time step, $x^o_{t+1}$, is again a piece-wise linear function of the control input $[\tilde{x}^r_{t+1}, \tilde{y}^r_{t+1}]$. As shown Fig. \ref{fig:friction_dynamics}a, the dynamics function consists of four pieces, where each piece corresponds to one contact mode. Note that the exact dynamics is flat for all $\tilde{y}^r_{t+1} > 0$, which provides no information about how the control affects the state. In contrast, the bundled dynamics (Fig. \ref{fig:friction_dynamics}c) has non-zero gradient even for $\tilde{y}^r_{t+1} > 0$.

For faster speed and better numerics, multi-body simulators (such as \cite{mujoco,Pang2020ACQ}) frequently relax the exact complementarity-based Coulomb friction constraints in \cite{anitescupotra, stewart2000rigid}. One such relaxation is proposed by Anitescu \cite{anitescu2006optimization}, which allows the forward dynamics to be solved as a quadratic program (QP) instead of a linear complementarity problem (LCP). Using Anitescu's relaxed constraints, the dynamics of $x^o_{t+1}$ with respect to $[\tilde{x}^r_{t+1}, \tilde{y}^r_{t+1}]$ (Fig. \ref{fig:friction_dynamics}b) can be shown to have a very similar structure as the exact Coulomb friction contact dynamics in Fig. \ref{fig:friction_dynamics}a. In particular, the relaxed dynamics is the same as the exact dynamics when the contact is sticking or separating. In sliding, a ``boundary layer" is introduced, in which the ball drags the box even when they are not in contact.

Although relaxed contact models such as \cite{anitescu2006optimization} sacrifice physical realism to a small extent, the bundled dynamics of the relaxed contact models behave similar to the bundled dynamics of the exact contact models, as shown in Fig. \ref{fig:friction_dynamics}c and \ref{fig:friction_dynamics}d. This suggests that algorithms using bundled dynamics for planning can safely capitalize the speed and efficiency of the relaxed contact models despite their mild physical inaccuracy. 
\end{example}

\subsection{Penalty-based Contacts}
\subsubsection{Description} In the penalty method \cite{huntcrossley}, the forces in the normal direction are approximated using a stiff spring. Denoting the signed distance between two objects as $\phi(q)$, we can write the normal-direction force with two modes of contact and no-contact.
\begin{equation}
    \small
    f_n = -k\min\{\phi(q),0\}.
\end{equation}

In the penalty approximation of Coulomb friction, sticking is often enforced by means of very viscous damping, while slipping is approximated by a constant multiple of the normal force. Let $\psi(q,\dot{q})$ be the relative tangential velocity between two objects at the point of contact.

\begin{equation}
    \small
    f_t = \mu f_n \qquad \mu = \begin{cases} c\psi(q,\dot{q}) & \; \text{ if } \; \psi(q,\dot{q}) \leq \; \psi_s \\ 
    \mu_d  & \; \text{ if } \; \psi(q,\dot{q}) > \psi_s \end{cases},
\end{equation}
where $\psi_s$ controls the threshold of stick and slip. Simple models of Coulomb friction sets $\psi_s$ to make the approximation continuous (i.e. $c\psi_s = \mu_d$), as done in \cite{add}. However, the Stribeck effect \cite{stribeck} tells us that dynamic friction is lesser in magnitude than static friction. Continuous approximations to the discontinuities of Stribeck effect is made in Drake \cite{drake}.

\subsubsection{Randomized Smoothing} 
\begin{figure}[h!]
\centering
\includegraphics[width = 0.40\textwidth]{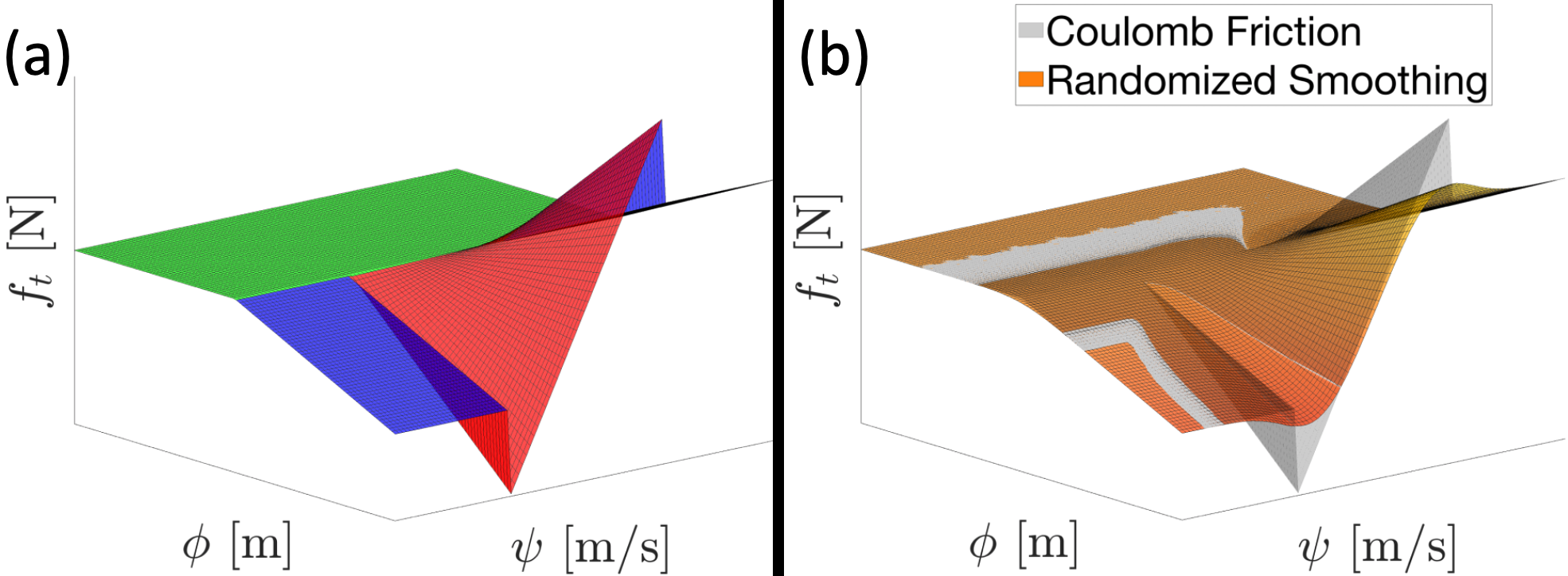}
\caption{(\textbf{a}) Penalty-based Coulomb friction model with Stribeck effect. Note the discontinuity between the blue (sliding) and red (sticking) pieces. (\textbf{b}) Discontinuity is removed by randomized smoothing.}
\label{fig:penalty_smoothing}
\vspace{-0.5cm}
\end{figure}

We apply randomized smoothing by sampling $\phi$ and $\psi$, and plot the results on Fig.\ref{fig:penalty_smoothing}. We note several effects of randomized smoothing: (\textbf{i}) in the normal direction, even if the objects are not in contact, repulsive forces are applied at a distance; (\textbf{ii}) tangential friction is also applied at a distance due to smoothing that occurs in the direction of $\phi$ in Fig.\ref{fig:penalty_smoothing}; (\textbf{iii}) discontinuous effects from the Stribeck effect is smoothed out.

\subsection{Key Insights}
\label{sec:contactsmoothinginsights}
\subsubsection{Problems with Exact Gradients} The non-smoothness of both the implicit time-stepping scheme and the penalty method suggests that the algorithms that rely on the local validity of linearization will suffer. The flatness of the gradients in non-penetrating configurations also suggests no direction of improvement.
\subsubsection{Relaxing Coulomb friction leads to better planning} Relaxations of the LCP version of Coulomb friction (e.g. \cite{anitescu2006optimization}) allow the application of sliding friction at a distance and coincidentally provide meaningful gradients useful for planning. For instance, the slope of sliding in Fig.\ref{fig:friction_dynamics}b suggests that the ball should be closer to the box in the direction of $y$ order to drag the box. This suggests that regularization used in popular simulators such as Mujoco \cite{mujoco} may fundamentally help planning algorithms achieve better performance. 
\subsubsection{Discontinuities of Contact Dynamics}
Implicit time-stepping and the penalty method both result in non-smooth, yet \textit{continuous} contact behavior with the exception of the Stribeck effect. The intuition behind the continuity of implicit time-stepping methods lies in the \textit{inelastic} approximation of contact behavior, while the continuity of penalty methods come from considering an infinitesimal section of time. 

However, we have yet to conclude the consequences of Theorem \ref{theorem:discontinuity} to be a minor nuisance. In more complex examples with non-smooth geometries, the normal of the object surface can also change in a discontinuous manner \cite{underactuated, Elandt2019APF}. In contrast, the zero-order bundled gradient can be free from the consequences of Theorem \ref{theorem:discontinuity}, which might suggest that the zero-order nature of popular RL algorithms such as the policy gradient \cite{policygradienttheorem} might coincidentally help in overcoming pitfalls of sampling gradients directly.

\begin{figure*}[thpb]
\vskip 0.05 true in
\centering\includegraphics[width = 0.9\textwidth]{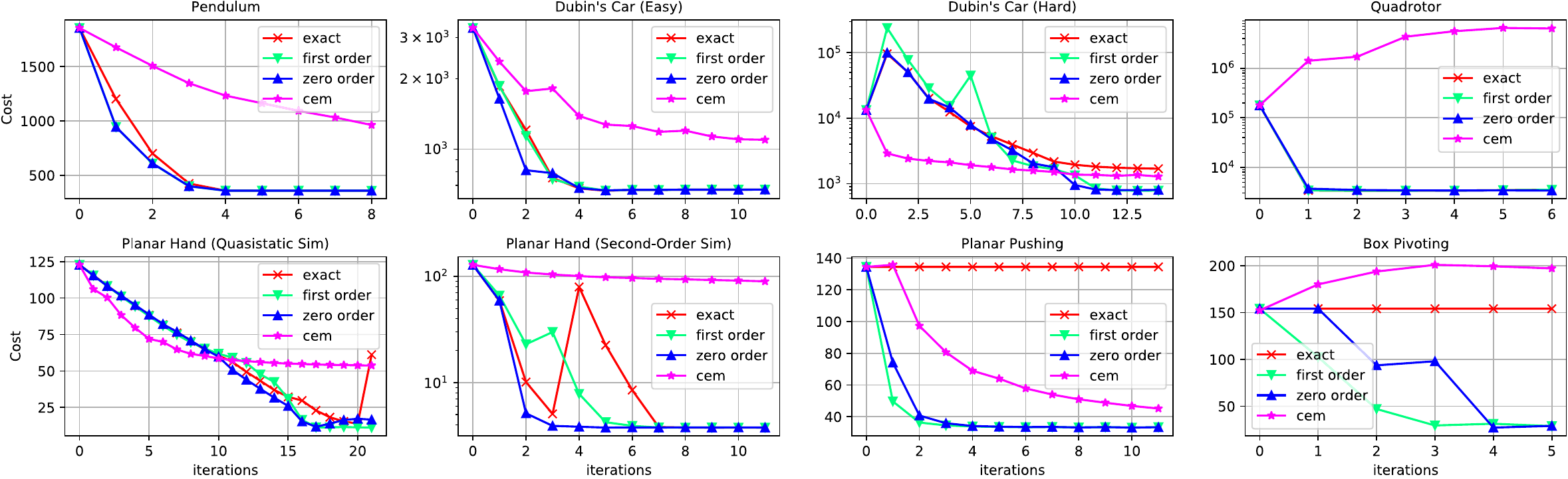}
\caption{Results for various systems on which iRS-MPC has been tested, with the objective of reaching some goal state. Lower cost indicates the system being closer to the goal configuration (in $L^2$ sense) at the end of the optimized trajectory. Top row: Smooth Systems. Bottom row: Systems with Contact. Results of the final iterations are illustrated in Fig.\ref{fig:banner}.}
\label{fig:results}
\vspace{-0.6cm}
\end{figure*}

\section{iterative Randomized Smoothing-MPC} 
The bundled gradient obtained via randomized smoothing can be used to replace the exact gradient in a variety of optimization problems, suggesting a new family of algorithms. In this work, we choose to replace the exact linearization with the bundled Jacobian in the trajectory optimization algorithm of iterative MPC (iMPC), which is an extension to the popular iLQR algorithm \cite{ddp,todorovilqr,tassa1} that supports constraints \cite[$\mathsection 10.7.2$]{underactuated}. We name our variant iterative Randomized Smoothing-MPC (iRS-MPC), which is similar to SaDDP \cite{saddp} or Unscented DDP \cite{plancher2017constrained}, but is uniquely applied in the presence of contact dynamics with the help of sampling.

The extension from iLQR to iMPC results in solving a QP of decreasing length at every time step. This is described by the subroutine:
\begin{equation}
    \small
    \begin{aligned}
        \text{MPC} \quad & (\bar{x}_j) = u^*_j, \text{ where } \\
        \min_{x_t,u_t}\quad & \bigg[\|x_T - x^d_T\|_{\mathbf{Q}_d}^2 + \sum_{t=j}^{T-1}\bigg(\|x_t - x^d_t\|_{\mathbf{Q}_t}^2 + \|u_t\|_{\mathbf{R}_t}^2\bigg)\bigg] \\
        \text{s.t.}\quad & x_{t+1} = \mathbf{A}_tx_t + \mathbf{B}_t u_t + c_t \quad \forall t\in\{j,\cdots,T-1\}, \\
        \quad & \mathbf{C}_t^x x_t  \leq d^x_t \quad \mathbf{C}^u_t u_t \leq d^u_t \qquad  \quad \forall t\in\{j,\cdots,T\}, \\
        \quad & x_j = \bar{x}_j.
    \end{aligned}
    \label{eq:mpc}
\end{equation}
Here, $\{\mathbf{C}^x_t,d^x_t\}$ are parameters of linear inequalities on state, and $\{\mathbf{C}^u_t,d^u_t\}$ are parameters of linear inequalities on inputs, and $\{x^d_t\}$ represents the desired trajectory. Note that the time indices depend on $j$; the horizon of the MPC decreases with our progress along the trajectory. Using this subroutine, we roll out the system using MPC at every timestep (whose horizons get shorter at every timestep) and iteratively repeat until a satisfactory solution has been reached (Alg.\ref{alg:irslqr}).

We introduce a variance scheduling function $\eta(\Sigma_0,k)$ which takes the initial variance $\Sigma_0$ and the current iteration number $k$ to produce a reduced variance $\Sigma_k$. Convergence to a local minima of the original problem is guaranteed as long as $\sum_k \eta(\Sigma_0,k)^2 < \infty$ \cite{duchi1, robbinsmonro, spsa}.
\begin{algorithm}[thpb]
\textbf{Given}: Initial state $x_0$, initial input trajectory $\bar{u}_t$\;
\textbf{Given}: Variance scheduler $\eta$, initialize $k=0$.\;
\textbf{Given}: Sampling distribution $\mu$ with variance $\Sigma_0$\;
$\{\bar{x}_t,\bar{u}_t\}^T_{t=0}\leftarrow$ Rollout $f$ from $x_0$ with $\bar{u}_t$\;
\While{Convergence}{
    \For{$0 \leq t < T$}{
    $\{\bar{\mathbf{A}}_t,\bar{\mathbf{B}}_t\}^T_{t=0}\leftarrow $ Use (\ref{eq:bundlejacobian}) or (\ref{eq:zeroorderjacobian}) around $(\bar{x}_t,\bar{u}_t)$ with variance $\Sigma_k=\eta(\Sigma_0, k)$\;
    $\bar{c}_t\leftarrow f(\bar{x}_t,\bar{u}_t) - \bar{\mathbf{A}}_t\bar{x}_t - \bar{\mathbf{B}}_t\bar{u}_t$\;
    }
    \For{$0 \leq t < T$}{
        Observe new nominal state $\bar{x}_t$\;
        Compute new nominal input $\bar{u}_t\leftarrow \text{MPC}(\bar{x}_t)$\;
        Apply $\bar{u}_t$ on the system. $\bar{x}_{t+1}=f(\bar{x}_t,\bar{u}_t)$
    }
    $k\leftarrow k+1$ 
 }
 \caption{iRS-MPC}
 \label{alg:irslqr}
\end{algorithm}

\section{Experiments and Results}
We show experimental comparisons of iRS-MPC (Alg. \ref{alg:irslqr}) against the baseline of using the same algorithm with exact gradients, which we refer to as iMPC. We also compare against the Cross-Entropy Method (CEM) \cite{cem}, a popular method for online trajectory optimization in model-based RL. These results are summarized in Fig. \ref{fig:results}.
\subsection{Simulation and Parallelization}
We test iRS-MPC on a quasi-dynamic simulation that relies on implicit time-stepping models of contact \cite{Pang2020ACQ}, as well as Drake \cite{drake}. We also parallelize iRS-MPC by having different threads per knot-point in $\{\bar{x}_t,\bar{u}_t\}^T_{t=0}$, which are responsible for sampling around the designated points. We use $N=100$ samples throughout all experiments to demonstrate the effectiveness of Monte-Carlo integration in higher dimensions. The computational cost of obtaining $\{\bar{\mathbf{A}}_t,\bar{\mathbf{B}}_t\}$ is equal to computing gradients of the dynamics $NT$ times, which can be reduced greatly with parallelization. The MPC computation requires solving $T$ number of QPs, each of decreasing length. Although more costly compared to iLQR, iMPC lets us handle constraints, and still fits well within the computational budget of modern QP solvers \cite{gurobi}. 
\subsection{Smooth Systems}
Will the bundled gradient be more effective than the gradient even for smooth systems? We investigate this question for three smooth systems: the pendulum, Dubin's car \cite{dubin}, and the quadrotor, which are described in \cite{underactuated}. In many cases, the performance of iRS-MPC is comparable to or better than that of exact linearization. However, when the initial guess supplied is not informative, the bundled gradient can power through local minima and still arrive at much better minima compared to exact gradients, as demonstrated by the example of Dubin's car.


\subsection{Non-Smooth Systems with Contact}
To test our hypothesis that the bundled gradient results in more stable behavior compared to using the exact gradient in the presence of contacts, we test iRS-MPC on three systems in robotic manipulation: a planar two-finger manipulation example with gravity, a planar pushing example \cite{francois}, and a box flipping (pivoting) example with gravity (Fig.\ref{fig:banner}). 

In the planar hand example, we start with an initial guess that makes contact. Although the exact gradient is able to drive the cost down, the fast changing gradients due to the switching of contact modes between iterations noticeably destabilize the algorithm. On the other hand, the first-order and zero-order bundled gradient are much more stable.

For the planar pushing and box pivoting example, we set the initial guess such that the robot is not in contact with the object. Due to flat gradients, iMPC is no longer able to make progress from this bad initialization. In contrast, iRS-MPC still succeeds in finding a valid descent direction and stably drives down the cost. These results demonstrate our insights in the problems of using exact gradients in Sec.\ref{sec:contactsmoothinginsights}.

We show that randomized smoothing is also able to stabilize second-order (fully-dynamic) systems with penalty-based contact models, whose gradients are obtained through Drake's autodiff feature. However, using the implicit quasi-dynamic formulation for planning resulted in much faster (x36) computation compared to the second-order penalty method due to less number of timesteps.
\vskip -0.1 true in
\subsection{First-order vs. Zero-Order bundled dynamics}
Throughout our work, we were constantly surprised to see that trajectory optimization with zero-order bundled dynamics was on par with, sometimes outperformed, the first-order one. Despite only having $100$ samples, the zero-order variant also produced good solutions in high-dimensional systems (e.g. quadrotor with $12$ states). which is consistent with the findings of \cite{saddp}. Combined with the robustness of zero-order optimization for sampling non-smooth functions (Example \ref{ex:heaviside}), this may suggest that zero-order smoothing of dynamics may be a promising alternative to first-order methods in tackling non-smooth contact problems.
\vskip -0.1 true in
\subsection{Comparison with Cross-Entropy Method}
We observed that the gradient-based iRS-MPC has faster convergence rate compared to CEM for all of our examples, and is also able to converge more tightly to local minima. In addition, for the examples of the quadrotor and box pivoting, CEM was not able to converge when the system was not open-loop stable. While CEM has its disadvantages, it performs better than the exact gradient in examples when the initial guess was not informative, such as box pushing. This supports our hypothesis that considering stochasticity may be key in explaining the success of RL methods.

\vspace{-0.1cm}
\section{Conclusion}
In this work, we have presented a method to obtain the bundled gradient through contact dynamics by using randomized smoothing. By applying randomized smoothing on two different contact models of implicit time-stepping and penalty, we answered what it means to take a stochastic formulation of contact dynamics. We then applied the bundled gradient to planning-through-contact problems, and showed that the bundled gradient enables much more stable convergence behavior compared to exact gradients. Through this, we answered how a stochastic formulation of dynamics might help the planning process. 

In addition to the qualitative analysis, our contribution includes taking the bundled gradient of implicit quasi-dynamic dynamics, whose explicit smooth approximations are difficult to make. Combining the stability of bundled gradients and the computational benefits of the quasi-dynamic simulation, we took a step towards stable and real-time solving of contact-rich manipulation problems.

\begingroup
\bibliographystyle{IEEETran}
{\small\bibliography{references}}

\begin{thebibliography}{10}
\providecommand{\url}[1]{#1}
\csname url@samestyle\endcsname
\providecommand{\newblock}{\relax}
\providecommand{\bibinfo}[2]{#2}
\providecommand{\BIBentrySTDinterwordspacing}{\spaceskip=0pt\relax}
\providecommand{\BIBentryALTinterwordstretchfactor}{4}
\providecommand{\BIBentryALTinterwordspacing}{\spaceskip=\fontdimen2\font plus
\BIBentryALTinterwordstretchfactor\fontdimen3\font minus
  \fontdimen4\font\relax}
\providecommand{\BIBforeignlanguage}[2]{{%
\expandafter\ifx\csname l@#1\endcsname\relax
\typeout{** WARNING: IEEEtran.bst: No hyphenation pattern has been}%
\typeout{** loaded for the language `#1'. Using the pattern for}%
\typeout{** the default language instead.}%
\else
\language=\csname l@#1\endcsname
\fi
#2}}
\providecommand{\BIBdecl}{\relax}
\BIBdecl

\bibitem{posa}
M.~Posa, C.~Cantu, and R.~Tedrake, ``A direct method for trajectory
  optimization of rigid bodies through contact,'' \emph{The International
  Journal of Robotics Research}, vol.~33, no.~1, pp. 69--81, 2014.

\bibitem{contactinvariantoptimization}
I.~Mordatch, E.~Todorov, and Z.~Popovi\'{c}, ``Discovery of complex behaviors
  through contact-invariant optimization,'' \emph{ACM Trans. Graph.}, vol.~31,
  no.~4, Jul. 2012.

\bibitem{add}
M.~Geilinger, D.~Hahn, J.~Zehnder, M.~B\"{a}cher, B.~Thomaszewski, and
  S.~Coros, ``Add: Analytically differentiable dynamics for multi-body systems
  with frictional contact,'' \emph{ACM Trans. Graph.}, vol.~39, no.~6, Nov.
  2020.

\bibitem{carpentierddp}
R.~Budhiraja, J.~Carpentier, C.~Mastalli, and N.~Mansard, ``Differential
  dynamic programming for multi-phase rigid contact dynamics,'' in \emph{2018
  IEEE-RAS 18th International Conference on Humanoid Robots (Humanoids)}, 2018,
  pp. 1--9.

\bibitem{francois}
F.~R. Hogan and A.~Rodriguez, \emph{Feedback Control of the Pusher-Slider
  System: A Story of Hybrid and Underactuated Contact Dynamics}.\hskip 1em plus
  0.5em minus 0.4em\relax Cham: Springer International Publishing, 2020, pp.
  800--815.

\bibitem{kent}
K.~Yunt and C.~Glocker, ``Trajectory optimization of mechanical hybrid systems
  using sumt,'' in \emph{9th IEEE International Workshop on Advanced Motion
  Control, 2006.}, 2006, pp. 665--671.

\bibitem{fu2016oneshot}
J.~Fu, S.~Levine, and P.~Abbeel, ``One-shot learning of manipulation skills
  with online dynamics adaptation and neural network priors,'' 2016.

\bibitem{nagabandi2019deep}
A.~Nagabandi, K.~Konoglie, S.~Levine, and V.~Kumar, ``Deep dynamics models for
  learning dexterous manipulation,'' 2019.

\bibitem{heess2017emergence}
N.~Heess, D.~TB, S.~Sriram, J.~Lemmon, J.~Merel, G.~Wayne, Y.~Tassa, T.~Erez,
  Z.~Wang, S.~M.~A. Eslami, M.~Riedmiller, and D.~Silver, ``Emergence of
  locomotion behaviours in rich environments,'' 2017.

\bibitem{florence2019selfsupervised}
P.~Florence, L.~Manuelli, and R.~Tedrake, ``Self-supervised correspondence in
  visuomotor policy learning,'' \emph{Robotics and Automation Letters}, 2020.

\bibitem{gradientsampling}
J.~Burke, F.~E. Curtis, A.~Lewis, M.~Overton, and L.~Simões, \emph{Gradient
  Sampling Methods for Nonsmooth Optimization}, 02 2020, pp. 201--225.

\bibitem{duchi1}
J.~Duchi, M.~Jordan, M.~Wainwright, and A.~Wibisono, ``Optimal rates for
  zero-order convex optimization: The power of two function evaluations,''
  \emph{IEEE Transactions on Information Theory}, vol.~61, 12 2015.

\bibitem{duchi2}
J.~Duchi, P.~Bartlett, and M.~Wainwright, ``Randomized smoothing for stochastic
  optimization,'' \emph{SIAM Journal on Optimization}, vol.~22, 03 2011.

\bibitem{yuxinchen}
\BIBentryALTinterwordspacing
Y.~Chen, ``Lecture notes for ele522: Large-scale optimization for data
  science,'' 2019. [Online]. Available:
  \url{https://www.princeton.edu/~yc5/ele522_optimization/}
\BIBentrySTDinterwordspacing

\bibitem{stewarttrinkle}
D.~Stewart and J.~J. Trinkle, ``An implicit time-stepping scheme for rigid body
  dynamics with coulomb friction.'' vol.~1, 01 2000, pp. 162--169.

\bibitem{anitescupotra}
M.~Anitescu and F.~Potra, ``Formulating dynamic multi-rigid-body contact
  problems with friction as solvable linear complementarity problems,''
  \emph{Nonlinear Dynamics}, vol.~14, 03 1997.

\bibitem{huntcrossley}
\BIBentryALTinterwordspacing
K.~H. Hunt and F.~R.~E. Crossley, ``{Coefficient of Restitution Interpreted as
  Damping in Vibroimpact},'' \emph{Journal of Applied Mechanics}, vol.~42,
  no.~2, pp. 440--445, 06 1975. [Online]. Available:
  \url{https://doi.org/10.1115/1.3423596}
\BIBentrySTDinterwordspacing

\bibitem{Pang2020ACQ}
T.~Pang and R.~Tedrake, ``A convex quasistatic time-stepping scheme for rigid
  multibody systems with contact and friction,'' in \emph{2021 IEEE
  International Conference on Robotics and Automation (ICRA)}.\hskip 1em plus
  0.5em minus 0.4em\relax IEEE, 2021.

\bibitem{brianmirtich}
B.~V. Mirtich, ``Impulse-based dynamic simulation of rigid body systems,''
  Ph.D. dissertation, 1996, aAI9723116.

\bibitem{tobia2}
T.~{Marcucci}, M.~{Gabiccini}, and A.~{Artoni}, ``A two-stage trajectory
  optimization strategy for articulated bodies with unscheduled contact
  sequences,'' \emph{IEEE Robotics and Automation Letters}, vol.~2, no.~1, pp.
  104--111, 2017.

\bibitem{terryfelix}
\BIBentryALTinterwordspacing
H.~T. Suh and Y.~Wang, ``Comparing effectiveness of relaxation methods for warm
  starting trajectory optimization through soft contact,'' 2019. [Online].
  Available:
  \url{http://people.csail.mit.edu/felixw/soft_contact/soft_contact.pdf}
\BIBentrySTDinterwordspacing

\bibitem{variablesmoothing}
A.~O. {Onol}, P.~{Long}, and T.~{Padır}, ``Contact-implicit trajectory
  optimization based on a variable smooth contact model and successive
  convexification,'' in \emph{ICRA}, 2019, pp. 2447--2453.

\bibitem{cottle2009linear}
R.~W. Cottle, J.-S. Pang, and R.~E. Stone, \emph{The linear complementarity
  problem}.\hskip 1em plus 0.5em minus 0.4em\relax SIAM, 2009.

\bibitem{anitescu2006optimization}
M.~Anitescu, ``Optimization-based simulation of nonsmooth rigid multibody
  dynamics,'' \emph{Mathematical Programming}, vol. 105, no.~1, pp. 113--143,
  2006.

\bibitem{convexsmoothcontactmodel}
E.~Todorov, ``A convex, smooth and invertible contact model for trajectory
  optimization,'' in \emph{ICRA}, 2011, pp. 1071--1076.

\bibitem{boot1963sensitivity}
J.~C. Boot, ``On sensitivity analysis in convex quadratic programming
  problems,'' \emph{Operations Research}, vol.~11, no.~5, pp. 771--786, 1963.

\bibitem{tobia}
T.~Marcucci and R.~Tedrake, ``Mixed-integer formulations for optimal control of
  piecewise-affine systems,'' in \emph{Proceedings of the 22nd ACM
  International Conference on Hybrid Systems: Computation and Control}, ser.
  HSCC ’19.\hskip 1em plus 0.5em minus 0.4em\relax New York, NY, USA:
  Association for Computing Machinery, 2019, p. 230–239.

\bibitem{scott}
Z.~Manchester, N.~Doshi, R.~J. Wood, and S.~Kuindersma, ``Contact-implicit
  trajectory optimization using variational integrators,'' \emph{The
  International Journal of Robotics Research}, vol.~38, no. 12-13, pp.
  1463--1476, 2019.

\bibitem{stochasticcomplementarity}
Y.~Tassa and E.~Todorov, ``Stochastic complementarity for local control of
  discontinuous dynamics,'' \emph{Robotics: Science and Systems}, 06 2010.

\bibitem{clarke}
F.~H. Clarke, \emph{Optimization and Nonsmooth Analysis}.\hskip 1em plus 0.5em
  minus 0.4em\relax Society for Industrial and Applied Mathematics, 1990.

\bibitem{subdifferentials}
J.~V. Burke, A.~S. Lewis, and M.~L. Overton, ``Approximating subdifferentials
  by random sampling of gradients,'' \emph{Mathematics of Operations Research},
  vol.~27, no.~3, pp. 567--584, 2002.

\bibitem{sqpgs}
F.~E. Curtis and M.~L. Overton, ``A sequential quadratic programming algorithm
  for nonconvex, nonsmooth constrained optimization,'' \emph{SIAM Journal on
  Optimization}, vol.~22, no.~2, pp. 474--500, 2012.

\bibitem{robbinsmonro}
H.~Robbins and S.~Monro, ``{A Stochastic Approximation Method},'' \emph{The
  Annals of Mathematical Statistics}, vol.~22, no.~3, pp. 400 -- 407, 1951.

\bibitem{kieferwolfowitz}
J.~Kiefer and J.~Wolfowitz, ``{Stochastic Estimation of the Maximum of a
  Regression Function},'' \emph{The Annals of Mathematical Statistics},
  vol.~23, no.~3, pp. 462 -- 466, 1952.

\bibitem{spsa}
J.~Spall, ``Multivariate stochastic approximation using a simultaneous
  perturbation gradient approximation,'' \emph{IEEE Transactions on Automatic
  Control}, vol.~37, no.~3, pp. 332--341, 1992.

\bibitem{gradientdominance}
H.~Karimi, J.~Nutini, and M.~Schmidt, ``Linear convergence of gradient and
  proximal-gradient methods under the polyak-{\l}ojasiewicz condition,'' in
  \emph{Machine Learning and Knowledge Discovery in Databases}.\hskip 1em plus
  0.5em minus 0.4em\relax Cham: Springer International Publishing, 2016, pp.
  795--811.

\bibitem{teg}
S.~P. Bangaru, J.~Michel, K.~Mu, G.~Bernstein, T.-M. Li, and J.~Ragan-Kelley,
  ``Systematically differentiating parametric discontinuities,'' \emph{ACM
  Trans. Graph.}, vol.~40, no.~4, Jul. 2021.

\bibitem{stewart2000rigid}
D.~E. Stewart, ``Rigid-body dynamics with friction and impact,'' \emph{SIAM
  review}, vol.~42, no.~1, pp. 3--39, 2000.

\bibitem{Landry2019BilevelOF}
B.~Landry, J.~Lorenzetti, Z.~Manchester, and M.~Pavone, ``Bilevel optimization
  for planning through contact: A semidirect method,'' \emph{International
  Symposium on Robotics Research (ISRR)}, vol. abs/1906.04292, 2019.

\bibitem{pang1996complementarity}
J.-S. Pang, J.~C. Trinkle, and G.~Lo, ``A complementarity approach to a
  quasistatic multi-rigid-body contact problem,'' \emph{Computational
  Optimization and Applications}, vol.~5, no.~2, pp. 139--154, 1996.

\bibitem{chavan2018stable}
N.~Chavan-Dafle and A.~Rodriguez, ``Stable prehensile pushing: In-hand
  manipulation with alternating sticking contacts,'' in \emph{ICRA}.\hskip 1em
  plus 0.5em minus 0.4em\relax IEEE, 2018, pp. 254--261.

\bibitem{mujoco}
E.~Todorov, T.~Erez, and Y.~Tassa, ``Mujoco: A physics engine for model-based
  control,'' in \emph{IROS}, 2012, pp. 5026--5033.

\bibitem{stribeck}
R.~Stribeck, \emph{Die wesentlichen Eigenschaften der Gleit- und Rollenlager},
  ser. Mitteilungen {\"u}ber Forschungsarbeiten auf dem Gebiete des
  Ingenieurwesens, insbesondere aus den Laboratorien der technischen
  Hochschulen.\hskip 1em plus 0.5em minus 0.4em\relax Julius Springer, 1903.

\bibitem{drake}
\BIBentryALTinterwordspacing
R.~Tedrake and the Drake Development~Team, ``Drake: Model-based design and
  verification for robotics,'' 2019. [Online]. Available:
  \url{https://drake.mit.edu}
\BIBentrySTDinterwordspacing

\bibitem{underactuated}
R.~Tedrake, ``Underactuated robotics: Algorithms for walking, running,
  swimming, flying, and manipulation (course notes for 6.832),'' 2021.

\bibitem{Elandt2019APF}
R.~Elandt, E.~Drumwright, M.~Sherman, and A.~Ruina, ``A pressure field model
  for fast, robust approximation of net contact force and moment between
  nominally rigid objects,'' \emph{IROS}, pp. 8238--8245, 2019.

\bibitem{policygradienttheorem}
R.~Sutton, D.~Mcallester, S.~Singh, and Y.~Mansour, ``Policy gradient methods
  for reinforcement learning with function approximation,'' \emph{Adv. Neural
  Inf. Process. Syst}, vol.~12, 02 2000.

\bibitem{ddp}
D.~H. JACOBSON and D.~Q. MAYNE, ``Differential dynamic programming. modern
  analytic and computational methods in science and mathematics, no. 24.
  american elsevier publ. co., inc., new york 1970. xvi, 208 s., 17 abb., dfl.
  51.50.'' \emph{Biometrische Zeitschrift}, vol.~15, no.~5, pp. 363--364, 1973.

\bibitem{todorovilqr}
W.~Li and E.~Todorov, ``Iterative linear quadratic regulator design for
  nonlinear biological movement systems.'' vol.~1, 01 2004, pp. 222--229.

\bibitem{tassa1}
Y.~Tassa, T.~Erez, and E.~Todorov, ``Synthesis and stabilization of complex
  behaviors through online trajectory optimization,'' in \emph{2012 IEEE/RSJ
  International Conference on Intelligent Robots and Systems}, 2012, pp.
  4906--4913.

\bibitem{saddp}
J.~Rajamäki, K.~Naderi, V.~Kyrki, and P.~Hämäläinen, ``Sampled differential
  dynamic programming,'' in \emph{2016 IEEE/RSJ International Conference on
  Intelligent Robots and Systems (IROS)}, 2016, pp. 1402--1409.

\bibitem{plancher2017constrained}
B.~Plancher, Z.~Manchester, and S.~Kuindersma, ``Constrained unscented dynamic
  programming,'' in \emph{IROS}.\hskip 1em plus 0.5em minus 0.4em\relax IEEE,
  2017, pp. 5674--5680.

\bibitem{cem}
R.~Rubinstein and D.~Kroese, \emph{The Cross-Entropy Method}.\hskip 1em plus
  0.5em minus 0.4em\relax Springer Information Science and Statistics, 2004.

\bibitem{gurobi}
{Gurobi Optimization, LLC}, ``{Gurobi Optimizer Reference Manual},'' 2021.

\bibitem{dubin}
L.~Dubins, ``On curves of minimal length with a constraint on average
  curvature, and with prescribed initial and terminal positions and tangents,''
  \emph{American Journal of Mathematics}, vol.~79, p. 497, 1957.

\end{thebibliography}
\endgroup
\end{document}